\documentclass{bmvc2k}

\usepackage{enumitem} 
\usepackage{multirow} 
\usepackage{makecell} 
\usepackage{pifont} 
\usepackage{amssymb} 
\usepackage{booktabs}
\usepackage{graphicx}

\title{Partial Convolution Meets Visual Attention}

\addauthor{Haiduo Huang}{huanghd@stu.xjtu.edu.cn}{2}
\addauthor{Fuwei Yang}{fuwei.yang@amd.com}{1}
\addauthor{Dong Li}{d.li@amd.com}{1}
\addauthor{Ji Liu}{liuji@amd.com}{1}
\addauthor{Lu Tian}{lu.tian@amd.com}{1}
\addauthor{Jinzhang Peng}{jinz.peng@amd.com}{1}
\addauthor{Pengju Ren}{pengjuren@xjtu.edu.cn}{2}
\addauthor{Emad Barsoum}{ebarsoum@amd.com}{1}

\addinstitution{
 Advanced Micro Devices, Inc.\\
 Beijing, China
}
\addinstitution{
 Xi'an Jiaotong University\\
 Xi'an, China
}

\runninghead{Student, Prof, Collaborator}{PATNet}

\def\eg{\emph{e.g}\bmvaOneDot}

\def\etc{etc.\@\xspace}
\def\ie{\emph{i.e.}\@\xspace}

\begin{document}

\maketitle

\begin{abstract}
  Designing an efficient and effective neural network has remained a prominent topic in computer vision research. Depthwise onvolution (DWConv) is widely used in efficient CNNs or ViTs, but it needs frequent memory access during inference, which leads to low throughput. FasterNet attempts to introduce partial convolution (PConv) as an alternative to DWConv but compromises the accuracy due to underutilized channels. To remedy this shortcoming and consider the redundancy between feature map channels, we introduce a novel {\bf P}artial visual {\bf AT}tention mechanism ({\bf PAT}) that can efficiently combine PConv with visual attention. Our exploration indicates that the partial attention mechanism can completely replace the full attention mechanism and reduce model parameters and FLOPs. Our PAT can derive three types of blocks: Partial Channel-Attention block (PAT\_ch), Partial Spatial-Attention block (PAT\_sp) and Partial Self-Attention block (PAT\_sf). First, PAT\_ch integrates the enhanced Gaussian channel attention mechanism to infuse global distribution information into the untouched channels of PConv. Second, we introduce the spatial-wise attention to the MLP layer to further improve model accuracy. Finally, we replace PAT\_ch in the last stage with the self-attention mechanism to extend the global receptive field. Building upon PAT, we propose a novel hybrid network family, named {\bf PATNet}, which achieves superior top-1 accuracy and inference speed compared to FasterNet on ImageNet-1K classification and excel in both detection and segmentation on the COCO dataset. Particularly, our PATNet-T2 achieves {\bf 1.3\%} higher accuracy than FasterNet-T2, while exhibiting {\bf 25\%} higher GPU throughput and {\bf 24\%} lower CPU latency.
\end{abstract}

\section{Introduction}
\label{sec:intro}

\begin{figure}[ht]
  \begin{tabular}{cc}
    \bmvaHangBox{\includegraphics[width=0.45\linewidth]{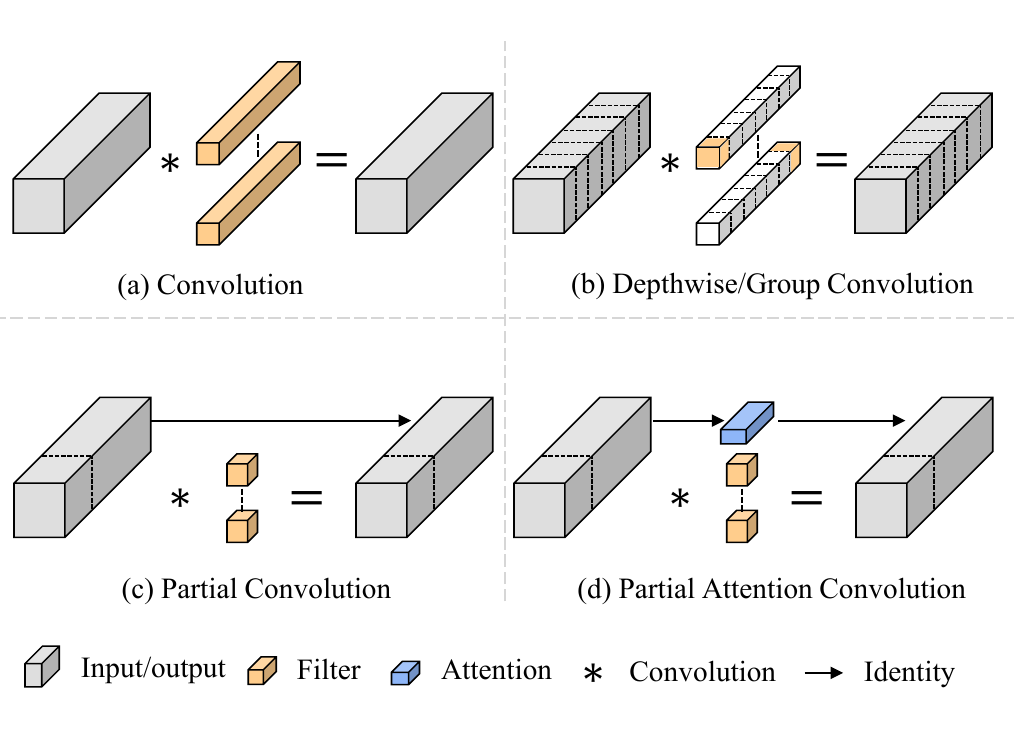}\label{fig:diff_modul}} &
    \bmvaHangBox{\includegraphics[width=0.45\linewidth]{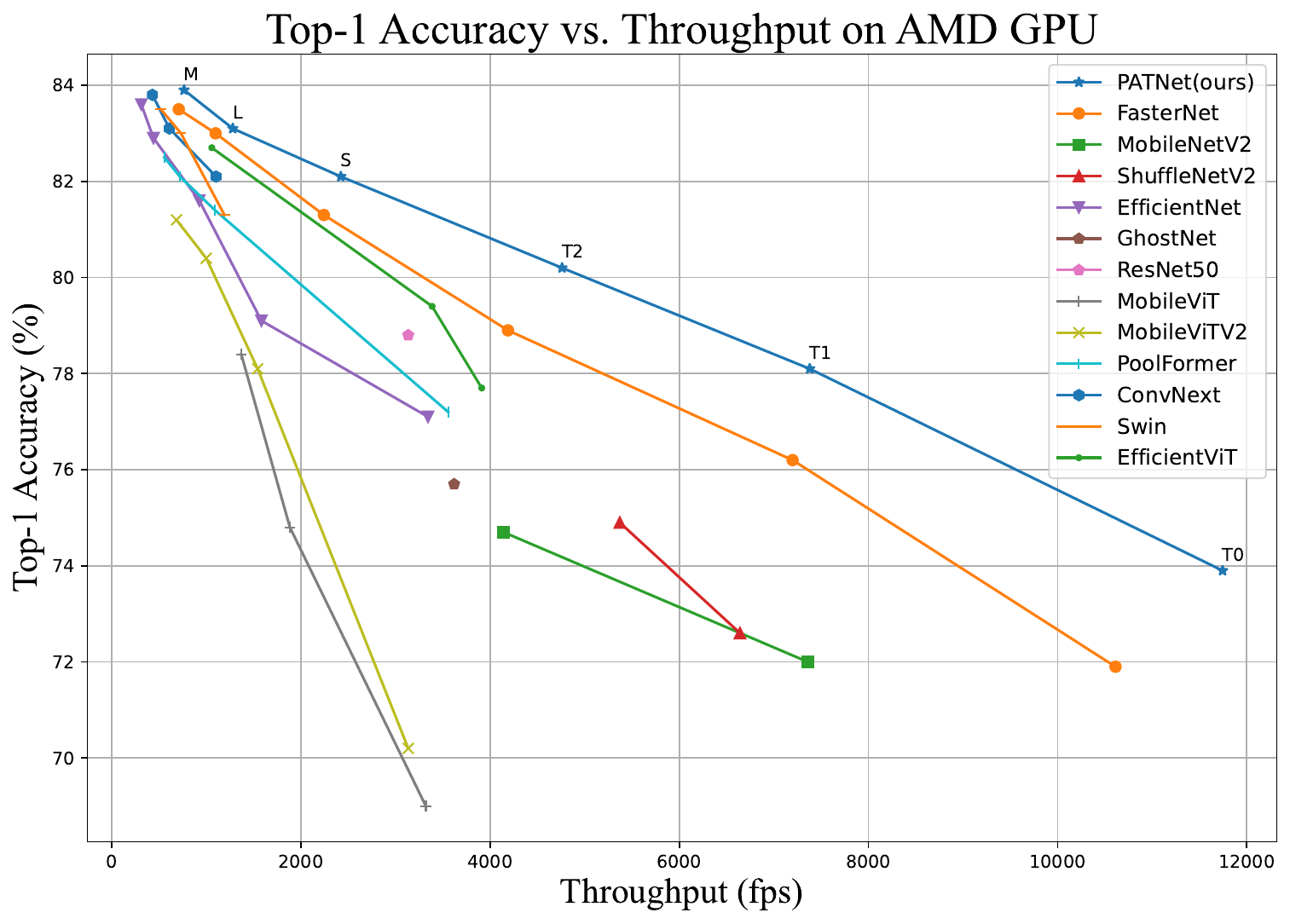}\label{fig:acc_throughput}} \\
    (a)                                                                                               & (b)
  \end{tabular}
  \caption{Comparison of different convolution types and efficient networks. Our PATNet incorporates the visual attention mechanism in Partial Convolution named Partial Attention Convolution, which surpasses the performance of FasterNet~\cite{Chen2023} on various model variants.}
\end{figure}

To design an efficient network, many prior works adopt depthwise separable convolution (DWConv)~\cite{Howard2017} as a substitute for regular dense convolution. For instance, some CNN-based models~\cite{Sandler2018a, Tan2019} leverage DWConv to reduce the model's FLOPs and parameters, while Hybrid-based models~\cite{yang2022focal, hou2022conv2former, Rao2022} employ DWConv to simulate self-attention operations to decrease computation complexity. Nevertheless, some studies~\cite{Ma2018, ding2022scaling} have revealed that DWConv may suffer from frequent memory access and low parallelism. Recent works have attempted to optimize the network's inference speed on specific hardware~\cite{Chen2023, chen2023vanillanet, Wang2023d, Vasu2023, Vasu2023a}. From the perspective of versatility, regular convolution (Conv) still has certain advantages.

Notably, FasterNet~\cite{Chen2023} proposes to use partial convolution (PConv) as an alternative to DWConv. Based on PConv, the FasterNet family achieves exceptional speed across various devices. PConv leverages redundancy within feature maps to selectively apply Conv to a subset of input channels, leaving the remaining channels untouched. That leads to lower FLOPs compared to regular Conv and higher FLOPS\footnote{FLOPs stands for floating-point operations, representing the number of arithmetic operations performed. FLOPS stands for floating-point operations per second, indicating the rate or speed at which these operations are executed within a given timeframe.} than DWConv~\cite{Chen2023}. However, we analyze that PConv underutilizes the untouched part and is constrained by the local dependencies inherent to CNNs, which may compromise accuracy. The primary reason for the decrease in accuracy is that PConv employs sparse (partial) parameters. So, how to maintain the inference speed of PConv while further enhancing its accuracy? Our motivation is integrating visual attention into partial convolution to enhance the feature representation ability of the untouched channels. We introduce a novel partial visual attention mechanism that can completely replace the conventional full attention mechanism without compromising accuracy and can reduce the model's parameter count and FLOPs compared with the full attention mechanism. The approach mainly involves substituting partial convolution with partial attention convolution, which is illustrated in Figure~\ref{fig:diff_modul}~(a).

How to choose a proper visual attention mechanism to achieve the optimal trade-off between model inference speed and accuracy? To address this problem, we propose three novel efficient partial visual attention blocks, \ie, Partial Channel-Attention block (PAT\_ch), Partial Spatial-Attention block (PAT\_sp) and Partial Self-Attention block (PAT\_sf). Firstly, we construct PAT\_ch by integrating an enhanced Gaussian channel attention mechanism~\cite{Hu2018}, facilitating richer inter-channel information interaction. Secondly, we extend the concept of partial convolution to MLP layer to further improve model performance. The convolution part of PA\_sp can be fused with the Conv1$\times$1 in the MLP during inference, resulting in efficient computation. Unlike previous spatial-wise attention~\cite{Woo2018}, our approach is simple and effective, involving only a Conv1$\times$1 operation and Hard-Sigmoid~\cite{Courbariaux2015} activation. Lastly, we refer to the MetaFormer-based~\cite{Yu2022a} paradigm and integrate global self-attention into the last stage of the CNN architecture to expand its global receptive field. The proposed PAT\_sf substantially boosts model accuracy in the ImageNet1k classification task.


In conclusion, the enhanced model is dubbed {\bf PATNet}, which achieves overall performance exceeding FasterNet in the ImageNet1K classification task while maintaining similar throughput, as is presented in Figure~\ref{fig:acc_throughput}~(b). Our main contributions can be described as:
\begin{itemize}
  \item We are the first to propose a novel partial visual attention mechanism that integrates visual attention into PConv, which can significantly improve model performance while minimizing the impact on inference speed.
  \item We develop three types of partial visual attention blocks including of PAT\_ch, PAT\_sp, and PAT\_sf. The PAT\_ch exhibits high potential as a replacement for regular convolution and DWConv. PAT\_sp can effectively reinforce MLP layers at minimal cost, while PAT\_sf integrates local and global features, achieving higher accuracy.
  \item Building upon PAT, we design a new hybrid-based model family named PATNet that shows improved performance on standard vision benchmarks over FasterNet with higher throughput and lower latency.
\end{itemize}

\section{Related Work}
\label{sec:related_work}
{~~~~~~\bf Efficient CNNs and ViTs.}
DWConv is widely adopted in the design of efficient neural networks, such as MobileNets~\cite{Sandler2018a, Howard2019}, EfficientNets~\cite{Tan2019, tan2021efficientnetv2}, MobileViT~\cite{Mehta2021}, and EdgeViT~\cite{Pan2022}. Despite its efficiency limitations on modern parallel devices, DWConv still holds unparalleled advantages on mobile devices. Given the drawbacks of DWConv, numerous works have aimed to improve it. For example,  RepLKNet~\cite{ding2022scaling} uses larger-kernel DWConv to alleviate the issue of underutilized calculations. PoolFormer~\cite{Yu2022a}, following the MetaFormer principles, achieves strong performance through spatial interaction with pooling operations alone. Recently, FasterNet~\cite{Chen2023} reduces FLOPs and memory accesses simultaneously by introducing partial convolution. Nevertheless, FasterNet does not outperform other vision models in accuracy. In contrast, our proposed PATNet addresses this limitation by integrating the visual attention mechanism into partial convolution, effectively enhancing the performance of FasterNet.

{\bf Attention Mechanism.}
Why are Vision Transformers (ViTs) so effective? Some studies attribute their success to the role of attention mechanisms~\cite{raghu2021vision, paul2022vision}. In visual tasks, attention mechanisms are commonly categorized into three types: Channel Attention, Spatial Attention, and Self-Attention. Some works~\cite{Mehta2022, Rao2022, Shaker2023, Cai2023b} employ various techniques to implement the Self-Attention mechanism efficiently, \eg, Linear Attention~\cite{Wang2020c, Cai2023b}. Furthermore, the effectiveness of Channel Attention and Spatial Attention has already been validated in SRM~\cite{Lee2019a}, SE-Net~\cite{Hu2018} and CBAM~\cite{Woo2018}. Similarly, we have incorporated attention mechanisms, but with a partial attention mechanism to mitigate the impact of element-wise multiplication on overall inference speed.

{\bf Additional Enhanced Technology} Some State-Of-The-Art networks employ additional technologies. For instance, MobileNetV3~\cite{Howard2019} utilizes NAS~\cite{Tan2019b} techniques to attain an optimal network structure. Networks like MobileOne~\cite{Vasu2023} and RIFormer~\cite{Wang2023d} rely on structured re-parameterization~\cite{Ding2021a} techniques, involving the addition of branches during training to expand its width and the merging of branches during inference to compress it. Furthermore, RIFormer~\cite{Wang2023d}, LeViT~\cite{Graham2021}, SwiftFormer~\cite{Shaker2023}, and RepViT~\cite{Wang2023b} leverage knowledge distillation~\cite{Huang2022} technology to transfer prior knowledge from large models to student models, thereby improving accuracy. Self-supervised pre-training~\cite{He2022} technology is employed in models like ConvNeXtV2~\cite{Woo2023} to achieve better model initialization. However, our PATNet follows regular training as the same FasterNet~\cite{Chen2023} without using bells and whistles.

\section{Methodology}
\label{sec:method}
In this section, we first elaborate on our motivation for integrating the visual attention mechanism into partial convolution and introduce {\bf P}artial visual {\bf At}tention mechanism ({\bf PAT}). Subsequently, we delve into our innovative Partial Channel-Attention block (PAT\_ch), Partial Spatial-Attention block (PAT\_sp), and Partial Self-Attention block (PAT\_sf). Finally, we design {\bf PATNet} architecture and explain its details.

\begin{figure}[ht]
  \centering
  \includegraphics[width=0.95\textwidth]{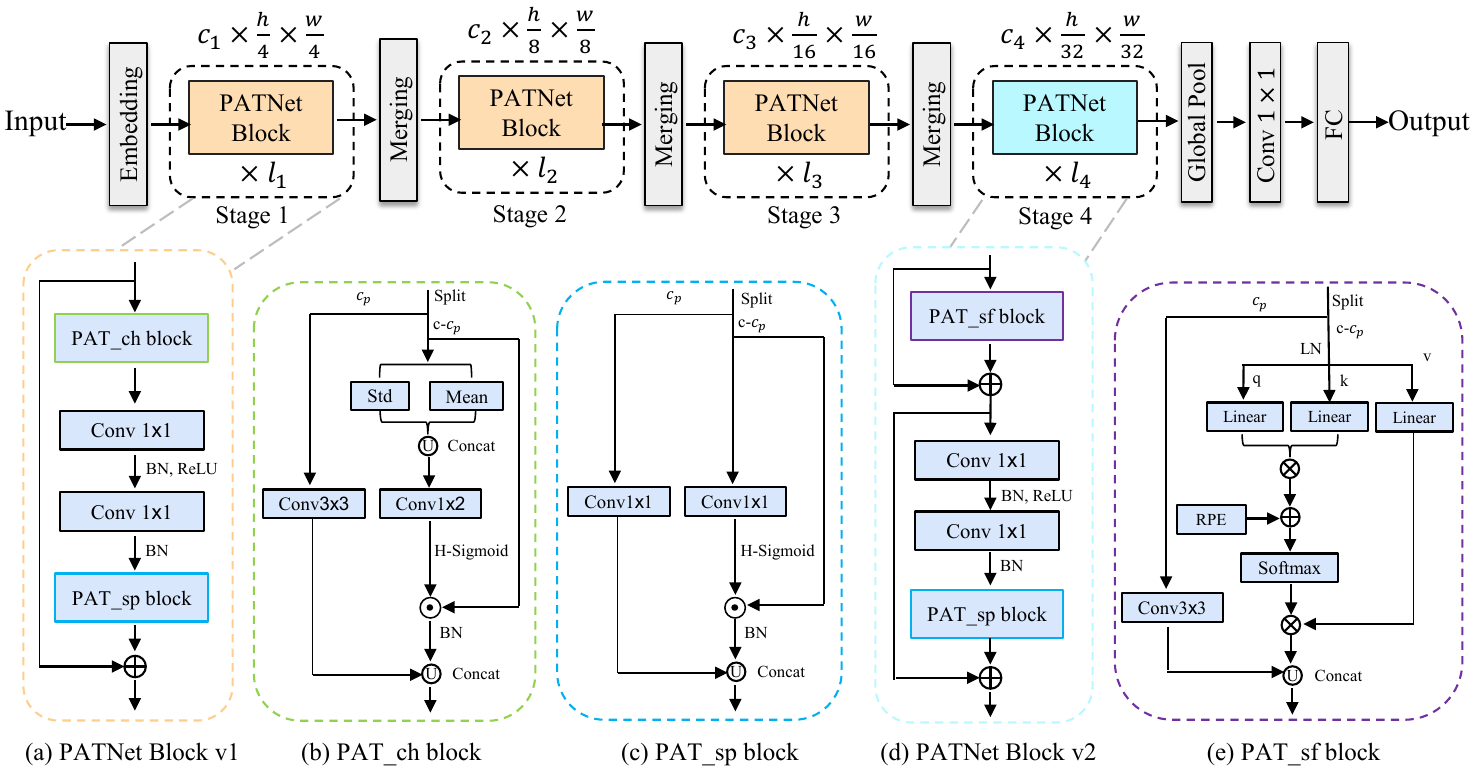}
  \caption{The overall architecture of our PATNet, consisting of four hierarchical stages, each incorporating a series of PATNet blocks followed by an embedding or merging layer. The last three layers are dedicated to feature classification. Where $\odot$ and $\otimes$ denote element-wise multiplication and matrix multiplication respectively.}
  \label{fig:network_overview}
\end{figure}

\subsection{Partial Visual Attention Mechanism}

Generally, designing an efficient and effective neural network necessitates comprehensive consideration and optimization from various perspectives, including fewer FLOPs, smaller model sizes, lower memory access, and comparable accuracy. Recently, the emerging FasterNet~\cite{Chen2023} may have met the aforementioned requirements to some degree and demonstrated its effectiveness across various vision tasks and terminal devices without additional technology enhancements. However, it does not exhibit a noticeable accuracy advantage when compared to models with similar parameters or FLOPs.

We empirically analyze that FasterNet mainly conducts Conv3$\times$3 operations on a portion of input channels of PConv, leaving the rest as direct identity mappings. These identity mappings are then concatenated with the processed Conv3$\times$3 portion. While this approach significantly reduces FLOPs and latency, it results in limited feature interaction and fusion, lacking global information interaction. Natural, we explore the integration of the visual attention mechanism into the identity mapping part (untouched part). Previous research~\cite {Han2020, Chen2023} has demonstrated that redundancy exists among feature map channels, making attention operations applied to the untouched parts a form of global information interaction.

Unlike regularly dense visual attention methods, our PAT is more efficient due to using only a subset of channels for the computationally expensive element-wise multiplication. Indeed, running two operations in parallel on separate branches allows for simultaneous computation, optimizing resource utilization on the GPU~\cite{kirk2016programming}. We also find that PAT is not only capable of applying channel-wise and spatial-wise mixing to enhance global information but also combines self-attention mechanisms to expand the model receptive field, proving to be highly effective. Below, We describe our PAT mechanism in formula.


Suppose the input and output of our PAT is $X,Y\in\mathbb{R}^{H\times W\times C}$, where $C$, $H$, $W$ represent the number of channels, height and width of a channel, respectively. We keep the number of channels unchanged after PAT. Then, the output can be formulated as
\begin{equation}
  Y =  Y^{C_p} \cup Y^{C-C_p} = Conv(X^{C_p}) \cup Atten(X^{C-C_p})
\end{equation}
where the symbol $\cup$ denotes the concatenation operation. $Conv$ denotes for regular convolution function and $Atten$ denotes for attention function, which can be one of channel attention, spatial attention and self attention. And $C_p=r_p \times C$ is defined as the number of front or last consecutive partial channels of the feature map. $r_p$  is a hyperparameter representing the $ratio$ used to select a portion of the channels. Detailed hyperparameter setup refer to the appendix.

\subsection{Efficient Integrated Visual Attention Information}
In this section, we explain our three types of partial visual attention in detail.

\begin{figure}
  \centering
  \includegraphics[width=0.95\textwidth]{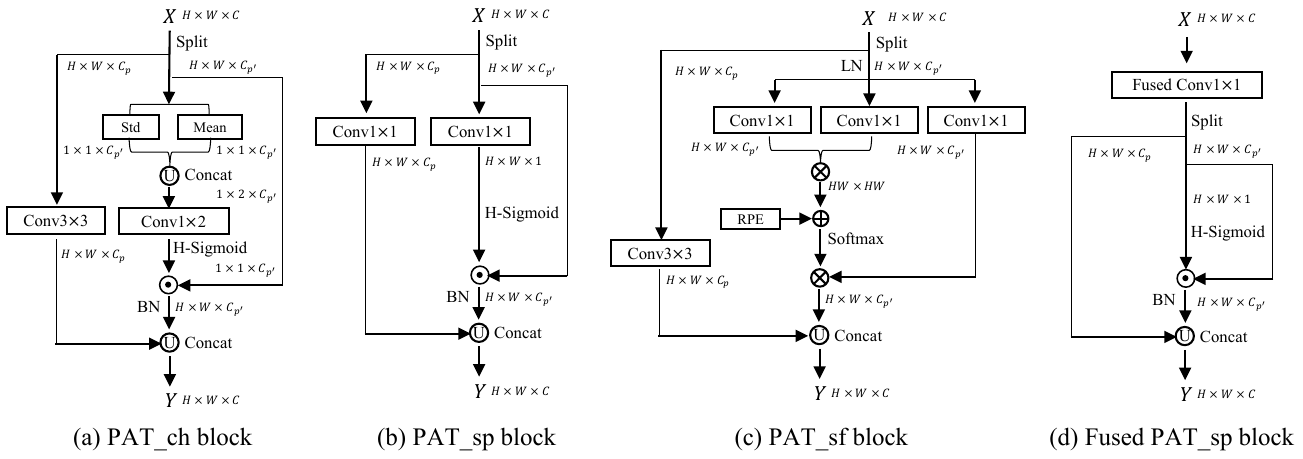}
  \caption{Combination of different partial visual attention blocks. Where $\odot$ and $\otimes$ denote element-wise multiplication and matrix multiplication respectively, and $C=C_p+C_{p^{'}}$.}
  \label{fig:attention_block}
\end{figure}

\textbf{PAT\_ch:} We integrate channel attention and Conv3$\times$3 because both involve spatial information interaction: Conv3x3 convolves and sums pixels within a local window, while our enhanced Gaussian-SE module computes channels' mean and variance to squeeze global spatial information. Unlike SENet~\cite{Hu2018}, it only considers the mean information of the channel and ignores the statistical information of std. Considering that the feature maps obey an approximately normal distribution~\cite{ioffe2015batch,Glorot2010} during training, we fully utilize the Gaussian statistical to express the channel-wise representation information, as shown in Figure~\ref{fig:attention_block}~(a).

\textbf{PAT\_sp:} We integrate spatial attention with Conv1$\times$1 because both operations mix channel wise information. Our spatial attention employs a point-wise convolution to squeeze global channel information into tensor with only 1 single channel. After passing through a Hard-Sigmoid activation, this tensor serves as the spatial attention map to weight features. We position PAT\_sp after the MLP layer, enabling the Conv1$\times$1 component of PAT\_sp to merge with the second Conv1$\times$1 in the MLP layer during inference, as shown in Figure~\ref{fig:attention_block}~(b) and Figure~\ref{fig:attention_block}~(d). This setup further minimizes the impact of attention on inference speed.

\textbf{PAT\_sf:} Since PAT\_sf also engages with spatial information interaction, it can replace PAT\_ch and extend the model's effective receptive field. However, because the computational complexity of self-attention operations increases quadratically with the size of the feature map, we restrict the use of PAT\_sf to the last stage to achieve a superior speed-accuracy trade-off. Beside, we employee relative position encoding (RPE)~\cite{Wu2021} into the attention map, which can further enhances model accuracy, as shown in Figure~\ref{fig:attention_block}~(c).

Notable, unlike conventional CNNs combined with attention, which process steps one after the other, we process steps simultaneously on the same input, improving the balance between speed and accuracy. In addition, our PAT is not limited to the above three combinations, it can be efficiently combined with more visual attention modules. Hence, the combination of the above three types of PAT blocks into a efficient PATNet.

\subsection{PATNet Architecture}

Our proposed PATNet refer to the recently introduced FasterNet~\cite{Chen2023}. The overall architecture, as depicted in Figure~\ref{fig:network_overview}, consists of four hierarchical stages, each of which precedes an embedding layer (a regular Conv4$\times$4 with stride 4) or a merging layer (a regular Conv2$\times$2 with stride 2). These layers serve for spatial downsampling and channel number expansion. Each stage comprises a set of PATNet blocks. In the first three stages of the PATNet, we employ "PATNet Block v1" including PAT\_ch block and PAT\_sp block, as shown in Figure~\ref{fig:network_overview} (a). However, we employ "PATNet Block v2" by replacing PAT\_ch with PAT\_sf in the last stage and modifying the shortcut connection way to achieve stable training, as shown in Figure~\ref{fig:network_overview} (d). Furthermore, we adjust the depth ratios across the four stages. In previous designs~\cite{Yu2022a,Liu2022b,Chen2023}, the depth of the last stage equals that of the first or second stage. We experimental find the critical importance of the last stage for network accuracy. Consequently, we adjusted the depth of the last stage to twice that of the first two stages. This adjustment substantially enhances model accuracy while minimally affecting throughput and latency. 

Following the FasterNet design principles, we maintain normalization or activation layers only after each intermediate Conv1$\times$1 to preserve feature diversity and achieve higher throughput. We also incorporate batch normalization into adjacent Conv layers to expedite inference without sacrificing performance. For the activation layer, the smaller PATNet variant uses GELU~\cite{hendrycks2016gaussian}, while the larger PATNet variant employs ReLU. Similarly, the last three layers consist of global average pooling, Conv1$\times$1, and a fully connected layer~\cite{Howard2019}. These layers collectively serve for feature transformation and classification. We offer tiny, small, medium, and large variants of PATNet, which are denoted as PATNet-T0/1/2, PATNet-S, PATNet-M, and PATNet-L. These variants share a similar architecture but differ in depth and width. The width of PATNet has been reduced compared to FasterNet to achieve faster inference speed. Detailed architectural specifications refer to the appendix.

\section{Experiments}
\label{sec:experiments}

\begin{table}[ht] \small
  \setlength\tabcolsep{3pt}   
  \renewcommand{\arraystretch}{0.8}  
  \centering
  \begin{tabular}{@{}lcccccccc@{}}
    \hline
    Network   &\makecell{Type}   & \makecell{Params\\(M)}	&\makecell{FLOPs\\(G)}	  &\makecell{Throughput\\V100 (FPS)$\uparrow$} &\makecell{Throughput\\MI250 (FPS)$\uparrow$}	&\makecell{Latency\\CPU (ms)$\downarrow$} &\makecell{Top-1\\(\%)$\uparrow$}  \\
    \hline
    ShuffleNetV2 x1.5\cite{Ma2018}      &cnn      & 3.5       & 0.30  &5315       & 6642        & 13.7        & 72.6       \\
    MobileNetV2\cite{Sandler2018a}      &cnn      & 3.5       & 0.31  &3924       & 7359        & 13.7        & 72.0       \\
    FasterNet-T0\cite{Chen2023}         &cnn      & 3.9       & 0.34  &{\bf 8546} & 10612       & {\bf 10.5}  & 71.9       \\
    MobileViTv2-0.5\cite{Mehta2022}     &hybrid   & 1.4       & 0.46  &3094       & 3135        & 15.8        & 70.2       \\
    PATNet-T0(ours)                     &hybrid   & 4.3       & 0.25  &7777       & {\bf 11744} & 12.2        & {\bf 73.9} \\
    \hline
    EfficientNet-B0\cite{Tan2019}       &cnn      & 5.3       & 0.39  &2934       & 3344        & 22.7        & 77.1       \\
    ShuffleNetV2 x2\cite{Ma2018}        &cnn      & 7.4       & 0.59  &4290       & 5371        & 22.6        & 74.9       \\
    MobileNetV2 x1.4\cite{Sandler2018a} &cnn      & 6.1       & 0.60  &2615       & 4142        & 21.7        & 74.7       \\
    FasterNet-T1\cite{Chen2023}         &cnn      & 7.6       & 0.85  &{\bf 4648} & 7198        & 22.2        & 76.2       \\
    PATNet-T1(ours)                     &hybrid   & 7.8       & 0.55  &4403       & {\bf 7379}  & {\bf 21.5}  & {\bf 78.1} \\
    \hline
    EfficientNet-B1\cite{Tan2019}       &cnn      & 7.8       & 0.70  &1730       & 1583        & 35.5        & 79.1       \\
    ResNet50\cite{He2015}               &cnn      & 25.6      & 4.11  &1258       & 3135        & 94.8        & 78.8       \\
    FasterNet-T2\cite{Chen2023}         &cnn      & 15.0      & 1.91  &2455       & 4189        & 43.7        & 78.9       \\
    PoolFormer-S12\cite{Yu2022a}        &hybrid   & 11.9      & 1.82  &1927       & 3558        & 56.1        & 77.2       \\
    MobileViTv2-1.0\cite{Mehta2022}     &hybrid   & 4.9       & 1.85  &1391       & 1543        & 41.5        & 78.1       \\
    EfficientViT-B1\cite{Cai2023b}      &hybrid   & 9.1       & 0.52  &3072       & 3387        & {\bf 25.7}  & 79.4       \\
    PATNet-T2(ours)                     &hybrid   & 12.6      & 1.03  &{\bf 3074} & {\bf 4761}  & 35.2        & {\bf 80.2} \\
    \hline
    EfficientNet-B3\cite{Tan2019}       &cnn      & 12.0      & 1.80  &768        & 926         & 73.5        & 81.6       \\
    ConvNeXt-T\cite{Liu2022b}           &cnn      & 28.6      & 4.47  &902        & 1103        & 99.4        & {\bf 82.1} \\
    FasterNet-S\cite{Chen2023}          &cnn      & 31.1      & 4.56  &1261       & 2243        & 96.0        & 81.3       \\
    PoolFormer-S36\cite{Yu2022a}        &hybrid   & 30.9      & 5.00  &675        & 1092        & 152.4       & 81.4       \\
    MobileViTv2-2.0\cite{Mehta2022}     &hybrid   & 18.5      & 7.50  &551        & 684         & 103.7       & 81.2       \\
    Swin-T\cite{Liu2021c}               &hybrid   & 28.3      & 4.51  &808        & 1192        & 107.1       & 81.3       \\
    PATNet-S(ours)                      &hybrid   & 29.0      & 2.71  &{\bf 1559} & {\bf 2422}  & {\bf 72.5}  & {\bf 82.1} \\
    \hline
    EfficientNet-B4\cite{Tan2019}       &cnn      & 19.0      & 4.20  &356        & 442         & 156.9       & 82.9       \\
    ConvNeXt-S\cite{Liu2022b}           &cnn      & 50.2      & 8.71  &510        & 610         & 185.5       & {\bf 83.1} \\
    FasterNet-M\cite{Chen2023}          &cnn      & 53.5      & 8.74  &621        & 1098        & 181.6       & 83.0       \\
    PoolFormer-M36\cite{Yu2022a}        &hybrid   & 56.2      & 8.80  &444        & 721         & 244.3       & 82.1       \\
    Swin-S\cite{Liu2021c}               &hybrid   & 49.6      & 8.77  &477        & 732         & 199.1       & 83.0       \\
    PATNet-M(ours)                      &hybrid   & 61.3      & 6.69  &{\bf 799}  & {\bf 1280}  & {\bf 155.3} & {\bf 83.1} \\
    \hline
    EfficientNet-B5\cite{Tan2019}       &cnn      & 30.0      & 9.90  &246        & 313         & 333.3       & 83.6       \\
    ConvNeXt-B\cite{Liu2022b}           &cnn      & 88.6      & 15.38 &322        & 430         & 317.1       & 83.8       \\
    FasterNet-L\cite{Chen2023}          &cnn      & 93.5      & 15.52 &384        & 709         & 312.5       & 83.5       \\
    PoolFormer-M48\cite{Yu2022a}        &hybrid   & 73.5      & 11.59 &335        & 556         & 322.3       & 82.5       \\
    Swin-B\cite{Liu2021c}               &hybrid   & 87.8      & 15.47 &315        & 520         & 333.8       & 83.5       \\
    PATNet-L(ours)                      &hybrid   & 104.3     & 11.91 &{\bf 426}  & {\bf 765}   & {\bf 272.5} & {\bf 83.9} \\
    \hline
  \end{tabular}
  \caption{Comparison on ImageNet-1k Benchmark: models with similar top-1 accuracy are grouped together. The best results are in bold. Full comparison please refer to appendix.}
  \label{tab:class_Benchmark}
\end{table}

\subsection{PATNet on ImageNet-1k Classification}

{\bf Setup.} ImageNet-1K~\cite{deng2009imagenet} is one of the most extensively used datasets in computer vision. It encompasses 1K common classes, consisting of approximately 1.3M training images and 50K validation images. We train our model on the ImageNet-1k dataset for 300 epochs using AdamW optimizer with 20 epochs linear warm-up. And we use the same regularization and augmentation techniques and multi-scale training as FasterNet \cite{Chen2023}. For detailed experimental settings, please refer to the appendix. In inference speed, we test the model's throughput in Nvidia V100 and AMD Instinct MI250 GPUs with batch size of 256, we test latency in AMD $\text{EPYC}^{TM}$ 73F3 CPU with one core.   

{\bf Results.} Table~\ref{tab:class_Benchmark} provides a comparison of our proposed PATNet models (T0, T1, T2, S, M, and L) with previous state-of-the-art cnn-based and hybrid-based models. The experimental results demonstrate that PATNet consistently surpasses recent models like FasterNet~\cite{Chen2023} across all model variants. For example, PATNet-T2 achieves 1.3\% higher accuracy than FasterNet-T2 while exhibiting around 25.2\%(or 13.7\%) increase in V100(or MI250) throughput and 24.1\% lower CPU latency. This comprehensive evaluation underscores the advantages of PATNet regarding accuracy and throughput (or latency) across various model sizes. So, it also demonstrates that the combination of visual attention and partial convolution significantly improves model performance without impacting throughput.

\subsection{PATNet on Downstream Tasks}
{\bf Setup.} We utilize the ImageNet1K pre-trained PATNet as the backbone within the Mask-RCNN~\cite{he2017mask} detector for object detection and instance segmentation on the MS-COCO 2017 dataset~\cite{lin2014microsoft}, comprising 118K training images and 5K validation images. To highlight the effectiveness of the backbone itself, we follow the FasterNet~\cite{Chen2023} approach and employ the AdamW~\cite{loshchilov2017decoupled} optimizer, conduct training of 12 epochs, use a batch size of 16, image size of 1333$\times$800, and maintain other training settings without further hyperparameter tuning.

\begin{table}[ht] \small
  \setlength\tabcolsep{3pt}   
  \renewcommand{\arraystretch}{0.8}  
  \centering
  \begin{tabular}{@{}lccccc cccc@{}}
    \hline
    Backbone              & \makecell{Params\\(M)}	&\makecell{FLOPs\\(G)} &\makecell{Throughput\\MI250 (FPS)$\uparrow$}	&\makecell{$AP^{b}\uparrow$}  &\makecell{$AP^{b}_{50}$} &\makecell{$AP^{b}_{75}$} &\makecell{$AP^{m}$} &\makecell{$AP^{m}_{50}$} &\makecell{$AP^{m}_{75}$}\\
    \hline
    ResNet50\cite{He2015}                 & 44.2    & 253   &121         & 38.0       & 58.6       & 41.4       & 34.4       & 55.1       & 36.7       \\
    PoolFormer-S24\cite{Yu2022a}          & 41.0    & 233   &68          & 40.1       & 62.2       & 43.4       & 37.0       & 59.1       & 39.6       \\
    PVT-Small x1.5\cite{Wang2021c}        & 44.1    & 238   &98          & 40.4       & 62.9       & 43.8       & 37.8       & 60.1       & 40.3       \\
    FasterNet-S\cite{Chen2023}            & 49.0    & 258   &121         & 39.9       & 61.2       & 43.6       & 36.9       & 58.1       & 39.7       \\
    PATNet-S(ours)                        & 46.9    & 216   &{\bf 122}   & {\bf 42.7} & {\bf 64.9} & {\bf 46.5} & {\bf 39.3} & {\bf 61.8} & {\bf 42.2} \\
    \hline
    ResNet101\cite{Mehta2021}             & 63.2    & 329   &62          & 40.4       & 61.1       & 44.2       & 36.4       & 57.7       & 38.8       \\
    ResNeXt101-32$\times$4d\cite{Xie2017} & 62.8    & 333   &51          & 41.9       & 62.5       & 45.9       & 37.5       & 59.4       & 40.2       \\
    PoolFormer-S36\cite{Yu2022a}          & 50.5    & 266   &44          & 41.0       & 63.1       & 44.8       & 37.7       & 60.1       & 40.0       \\
    PVT-Medium\cite{Wang2021c}            & 63.9    & 295   &52          & 42.0       & 64.4       & 45.6       & 39.0       & 61.6       & 42.1       \\
    FasterNet-M\cite{Chen2023}            & 71.2    & 344   &62          & 43.0       & 64.4       & 47.4       & 39.1       & 61.5       & 42.3       \\
    PATNet-M(ours)                        & 78.2    & 295   &{\bf 65}    & {\bf 44.3} & {\bf 65.8} & {\bf 48.5} & {\bf 40.6} & {\bf 63.3} & {\bf 43.7} \\
    \hline
    ResNeXt101-64$\times$4d\cite{Xie2017} & 101.9   & 487   &29          & 42.8       & 63.8       & 47.3       & 38.4       & 60.6       & 41.3       \\
    PVT-Large$\times$4d\cite{Wang2021c}   & 81.0    & 358   &26          & 42.9       & 65.0       & 46.6       & 39.5       & 61.9       & 42.5       \\
    FasterNet-L\cite{Chen2023}            & 110.9   & 484   &35          & 44.0       & 65.6       & 48.2       & 39.9       & 62.3       & 43.0       \\
    PATNet-L(ours)                        & 122.0   & 397   &{\bf 39}    & {\bf 44.7} & {\bf 66.3} & {\bf 49.0} & {\bf 41.0} & {\bf 63.7} & {\bf 44.2} \\
    \hline
  \end{tabular}
  \caption{Results using PATNet as a backbone on dense prediction tasks: Object detection and instance segmentation benchmark on the COCO dataset.}
  \label{tab:dect_and_seg_Benchmark}
\end{table}

{\bf Results.} Table~\ref{tab:dect_and_seg_Benchmark} presents a comparison of PATNet with representative models, reporting performance in terms of average precision (mAP) for both detection and instance segmentation. As shown in Table~\ref{tab:dect_and_seg_Benchmark}, PATNet consistently outperforms FasterNet, achieving higher average precision (AP) while maintaining similar latency. The results further confirm the generalization capabilities of our proposed PATNet across various tasks.

\subsection{Ablation Studies}
\label{sec:ablation}
\textbf{Partial Attention vs. Full Attention.}
\label{sec:pa_vs_fa}
To prove the superiority of our PAT over full attention mechanisms, we conduct comparative experiments on the PATNet\_T2, as shown in Table~\ref{tab:abla_block_acc}. Specifically, We replace PAT blocks with corresponding full visual attention for comparison respectively. Full visual attention involves conducting visual attention calculations on all channels of the input feature map, without considering the split operation and convolution operation of another branch, which is the common way of conventional visual attention mechanism. The results demonstrate the feasibility of performing attention operations on part channels and also confirm the effectiveness of our improved visual attention mechanism. The results indicate that our PAT achieves a superior balance between inference speed and performance compared to the Full visual attention counterpart.
\begin{table}[ht]
  \setlength\tabcolsep{3.8pt}   
   \renewcommand{\arraystretch}{0.8}  
  \centering
  \begin{tabular}{@{}lcccc ccc@{}}
    \hline
    \makecell{ch} & \makecell{sp} & \makecell{sf} & \makecell{Params (M)} & \makecell{FLOPs (G)} & \makecell{Throughput (FPS)$\uparrow$} & \makecell{Latency (ms)$\downarrow$} & \makecell{Top-1 (\%)$\uparrow$} \\
    \hline
    P             & P    & P     & 12.6       & 1.03     & {\bf 4761}   & {\bf 35.2}  & {\bf 80.2}                          \\
    F             & P    & P     & 13.0       & 1.04     & 4662         & 36.5        & 80.1                          \\
    P             & F    & P     & 12.6       & 1.04     & 4688         & 35.6        & 79.9                          \\
    P             & P    & F     & 14.5       & 1.12     & 4600         & 38.6        & {\bf 80.2}                         \\
    \hline
  \end{tabular}
  \caption{Comparison on PATNet-T2 of partial attention (P), and full attention (F) on ImageNet1K dataset. Where the "ch", "sp", and "sf" denote channel-wise attention, spatial-wise attention, and self-attention respectively.}
  \label{tab:abla_block_acc}
\end{table}

\textbf{Effect of PAT blocks.}
To demonstrate the individual effects of our three PAT blocks, we conducted ablation studies by progressively adding each PAT block one by one, as indicated in Table~\ref{tab:abla_PAT_acc}. Experiment results indicate that the three proposed PAT blocks consistently enhance model performance.
\begin{table}[ht]
  \setlength\tabcolsep{4.0pt}   
  \centering
  \begin{tabular}{@{}lcccc cccc@{}}
    \hline
    \makecell{Stages} & \makecell{PAT\_ch} & \makecell{PAT\_sp} & \makecell{PAT\_sf} & \makecell{Params\\(M)} & \makecell{FLOPs\\(G)} & \makecell{Throughput\\(FPS)$\uparrow$} & \makecell{Latency\\(ms)$\downarrow$} & \makecell{Top-1\\(\%)$\uparrow$} \\
    \hline
    2-2-6-4           &                    &              &              &11.1  &0.92      & {\bf 6405} & {\bf 25.7} & 76.0                          \\
    2-2-6-4           & \ding{51}          &              &              &11.1  &0.92      & 5440       & 30.9       & 77.4                          \\
    2-2-6-4           & \ding{51}          & \ding{51}    &              &11.5  &0.92      & 5157       & 31.7       & 78.9                          \\
    2-2-6-4           & \ding{51}          & \ding{51}    & \ding{51}    &12.6  &1.03      & 4761       & 35.2       & {\bf 80.2}                    \\
    2-2-8-2           & \ding{51}          & \ding{51}    & \ding{51}    &9.7   &0.98      & 4976       & 32.7       & 78.8                          \\
    \hline
  \end{tabular}
  \caption{Ablation experiments of PATNet-T2 with different configurations of PAT blocks across different model stages on the ImageNet1K dataset.}
  \label{tab:abla_PAT_acc}
\end{table}

\textbf{Different Stage Settings.}
We adhere to the model design convention of utilizing four stages. However, previous works overlook the importance of the last stage, \eg, FasterNet~\cite{Chen2023} and MetaFormer ~\cite{Yu2022a}. We conduct the comparative experiments between different stage settings (2-2-6-4 vs. 2-2-8-2). The last two rows of Table ~\ref{tab:abla_PAT_acc} show that our adjusted stage depths (i.e., 2-2-6-4) can bring more accuracy gain (78.8\% $\to$ 80.2\%) with a slight performance drop.

\textbf{Partial Visual Convolution vs. Regular (or DepthWise) Convolution.}
To further verify the advantages of our proposed partial visual convolution (PAT\_ch) over regular convolution (Conv), we conducted ablation experiments on PATNet-T2 in Table~\ref{tab:abla_PAT_ch}. To make a fair comparison, we widen DWConv to keep the throughput of the three convolution types in the same range. Experimental results show that our proposed PAT\_ch surpasses regular (or DepthWise) convolution in all metrics including Params, Flops, throughput, latency and Top-1 accuracy, which validates the efficiency and effectiveness of PAT. 

\begin{table}[ht]
  \setlength\tabcolsep{4.0pt}   
  \centering
  \begin{tabular}{@{}lcccc cc@{}}
    \hline
    \makecell{Conv3$\times$3} & \makecell{Params (M)} & \makecell{FLOPs (G)} & \makecell{Throughput (FPS)$\uparrow$} & \makecell{Latency (ms)$\downarrow$} & \makecell{Top-1 (\%)$\uparrow$} \\
    \hline
    PAT\_ch        & 12.6             & 1.03           & 4761               & 35.2               & {\bf 80.2}                          \\
    Conv           & 15.8             & 2.12           & 4190               & 49.9               & 79.9                          \\
    DWConv         & 15.8             & 1.28           & 4017               & 35.4               & 79.6                          \\
    \hline
  \end{tabular}
  \caption{Ablation on PATNet-T2 with different convolution types on ImageNet.}
  \label{tab:abla_PAT_ch}
\end{table}

\section{Conclusion}
\label{sec:conclusion}
This paper introduces the concept of partial visual attention mechanism which strategically integrates visual attention mechanisms into partial convolution. We propose three novel partial visual attention blocks including of Partial Channel-Attention block, Partial Spatial-Attention block, and Partial Self-Attention block, which enable models to achieve higher performance while maintaining efficiency. Building upon these innovations, we introduce the PATNet network which outperforms the recent FasterNet network in ImageNet1K classification, as well as COCO detection and segmentation tasks. This underscores the effectiveness of the Partial visual Attention mechanism and signifies a novel convolution approach that strikes an optimal balance between high accuracy and efficiency for various vision tasks. The idea of partial attention still has great potential in the natural language processing (NLP) or large language model (LLM) domains.

\appendix
\section{Appendix}
\subsection{Overview}
In this supplementary material, we present more explanations and experimental results.
\begin{itemize}
  \item We first make detailed explanations of our experimental setting and different PATNet variants.
  \item We then present a full comparison on ImageNet-1k Benchmark.
  \item We also provide further ablation studies for our proposed Partial Visual Attention mechanism ({\bf PAT}).
\end{itemize}

\subsection{Clarifications on Experimental Setting}
Firstly, the configurations of different PATNet variants are presented in Table~\ref{tab:configuration}. We also provide ImageNet-1k training and evaluation settings in Table~\ref{tab:imagenet_settings}. They can be used for reproducing our main results in Figure~1 of the main paper. Different PATNet variants vary in the magnitude of regularization and augmentation techniques. The magnitude increases as the model becomes larger to alleviate overfitting and improve accuracy. Note that most of the compared works in~Figure 1 of the main paper, \eg, MobileViT, FastNet, ConvNeXt, Swin, \etc, also adopt such advanced training techniques (ADT). Some even heavily rely on the hyper-parameter search. For others w/o ADT, \eg, ShuffleNetV2, MobileNetV2, and GhostNet, though the comparison is not totally fair, we include them for reference. 

\begin{table*}[ht]
  \small
  \renewcommand{\arraystretch}{0.8}  
  \centering
  \resizebox{.98\linewidth}{!}{
    \begin{tabular}{@{}c|c|c|c|c|c|c|c|c|c@{}}  
      \toprule
      Name                            & Output size                                & \multicolumn{2}{c|}{Layer specification}                                                                                                                                                & T0               & T1   & T2   & S     & M    & L           \\
      \midrule
      Embedding                       & \large{$\frac{h}{4} \times \frac{w}{4}$}   & \begin{tabular}[c]{@{}c@{}}Conv\_4\_$c$\_4,\\ BN\end{tabular}                                                                                                                           & \# Channels $c$  & 32   & 48   & 64    & 96   & 128  & 160  \\
      \midrule
      Stage 1                         & \large{$\frac{h}{4} \times \frac{w}{4}$}   & $\left[ \text{\begin{tabular}[c]{@{}c@{}}PAT\_ch\_3\_$c$\_1\_1/4,\\ Conv\_1\_$2c$\_1,\\ BN, Acti,\\ Conv\_1\_$c$\_1,\\ PAT\_sp\_1\_$c$\_1\_1/4 \end{tabular}}  \right] \times b_1 $     & \# Blocks $b_1$  & 1    & 2    & 2     & 2    & 2    & 2    \\
      \midrule
      Merging                         & \large{$\frac{h}{8} \times \frac{w}{8}$}   & \begin{tabular}[c]{@{}c@{}}Conv\_2\_$2c$\_2,\\ BN\end{tabular}                                                                                                                          & \# Channels $2c$ & 64   & 96   & 128   & 192  & 256  & 320  \\
      \midrule
      Stage 2                         & \large{$\frac{h}{8} \times \frac{w}{8}$}   & $\left[ \text{\begin{tabular}[c]{@{}c@{}}PAT\_ch\_3\_$2c$\_1\_1/4,\\ Conv\_1\_$4c$\_1,\\ BN, Acti,\\ Conv\_1\_$2c$\_1,\\ PAT\_sp\_1\_$2c$\_1\_1/4 \end{tabular}}  \right] \times b_2 $  & \# Blocks $b_2$  & 2    & 2    & 2     & 2    & 3    & 3    \\
      \midrule
      Merging                         & \large{$\frac{h}{16} \times \frac{w}{16}$} & \begin{tabular}[c]{@{}c@{}}Conv\_2\_$4c$\_2,\\ BN\end{tabular}                                                                                                                          & \# Channels $4c$ & 128  & 192  & 256   & 384  & 512  & 640  \\
      \midrule
      Stage 3                         & \large{$\frac{h}{16} \times \frac{w}{16}$} & $\left[ \text{\begin{tabular}[c]{@{}c@{}}PAT\_ch\_3\_$4c$\_1\_1/4,\\ Conv\_1\_$8c$\_1,\\ BN, Acti,\\ Conv\_1\_$4c$\_1,\\ PAT\_sp\_1\_$4c$\_1\_1/4 \end{tabular}}  \right] \times b_3 $  & \# Blocks $b_3$  & 6    & 6    & 6     & 9    & 16   & 20   \\
      \midrule
      Merging                         & \large{$\frac{h}{32} \times \frac{w}{32}$} & \begin{tabular}[c]{@{}c@{}}Conv\_2\_$8c$\_2,\\ BN\end{tabular}                                                                                                                          & \# Channels $8c$ & 256  & 384  & 512   & 768  & 1024 & 1280 \\
      \midrule
      Stage 4                         & \large{$\frac{h}{32} \times \frac{w}{32}$} & $\left[ \text{\begin{tabular}[c]{@{}c@{}}PAT\_ch\_3\_$8c$\_1\_1/4,\\ Conv\_1\_$16c$\_1,\\ BN, Acti,\\ Conv\_1\_$8c$\_1,\\ PAT\_sf\_1\_$8c$\_1\_1/4 \end{tabular}}  \right] \times b_4 $ & \# Blocks $b_4$  & 4    & 4    & 4     & 4    & 4    & 4    \\
      \midrule
      Classifier                      & $1  \times 1$                              & \begin{tabular}[c]{@{}c@{}}Global average pool,\\ Conv\_1\_1280\_1,\\ Acti,\\ FC\_1000\end{tabular}                                                                                     & Acti             & GELU & GELU & ReLU  & ReLU & ReLU & ReLU \\
      \midrule
      \multicolumn{4}{c|}{Params (M)} & 4.3                                        & 7.8                                                                                                                                                                                     & 12.6             & 29.0 & 61.3 & 104.4                      \\
      \midrule
      \multicolumn{4}{c|}{FLOPs (G)}  & 0.25                                       & 0.55                                                                                                                                                                                    & 1.03             & 2.71 & 6.69 & 11.91                       \\
      \bottomrule
    \end{tabular}%
  }
  \caption{Configurations of different PATNet variants. ``Conv\_$k$\_$c$\_$s$'' means a convolutional layer with the kernel size of $k$, the output channels of $c$, and the stride of $s$. ``PAT\_ch$\_k\_c\_s\_r$'' means a partial convolution with an extra parameter, the partial ratio of $r$. ``FC\_1000'' means a fully connected layer with 1000 output channels. $h \times w$ is the input size while $b_i$ is the number of PATNet blocks at stage $i$. The FLOPs are calculated given the input size of $224 \times 224$.}
  \label{tab:configuration}
\end{table*}

\begin{table*}[ht]
  \small
  \renewcommand{\arraystretch}{0.7}  
  \centering
  \resizebox{0.83\linewidth}{!}{
    \begin{tabular}{@{}l|cccccc@{}}
      \toprule
      Variants           & T0                                                                 & T1                               & T2                               & S                                & M     & L     \\
      \midrule
      Train Res          & \multicolumn{6}{c}{Random select from \{128,160,192,224,256,288\}}                                                                                                                          \\
      Test Res           & \multicolumn{6}{c}{224}                                                                                                                                                                     \\
      \midrule
      Epochs             & \multicolumn{6}{c}{300}                                                                                                                                                                     \\
      \# of forward pass & \multicolumn{6}{c}{188k}                                                                                                                                                                    \\
      \midrule
      Batch size         & 4096                                                               & 4096                             & 4096                             & 4096                             & 2048  & 2048  \\
      Optimizer          & \multicolumn{6}{c}{AdamW}                                                                                                                                                                   \\
      Momentum           & \multicolumn{6}{c}{0.9/0.999}                                                                                                                                                               \\
      LR                 & 0.004                                                              & 0.004                            & 0.004                            & 0.004                            & 0.002 & 0.002 \\
      LR decay           & \multicolumn{6}{c}{cosine}                                                                                                                                                                  \\
      Weight decay       & 0.005                                                              & 0.01                             & 0.02                             & 0.03                             & 0.05  & 0.05  \\
      Warmup epochs      & \multicolumn{6}{c}{20}                                                                                                                                                                      \\
      Warmup schedule    & \multicolumn{6}{c}{linear}                                                                                                                                                                  \\
      \midrule
      Label smoothing    & \multicolumn{6}{c}{0.1}                                                                                                                                                                     \\
      Dropout            & \multicolumn{6}{c}{{\color[HTML]{9B9B9B} \ding{55}}}                                                                                                                                        \\
      Stoch. Depth       & {\color[HTML]{9B9B9B} \ding{55}}                                   & 0.02                             & 0.05                             & 0.1                              & 0.2   & 0.3   \\
      Repeated Aug       & \multicolumn{6}{c}{{\color[HTML]{9B9B9B} \ding{55}}}                                                                                                                                        \\
      Gradient Clip.     & {\color[HTML]{9B9B9B} \ding{55}}                                   & {\color[HTML]{9B9B9B} \ding{55}} & {\color[HTML]{9B9B9B} \ding{55}} & {\color[HTML]{9B9B9B} \ding{55}} & 1     & 0.01  \\
      \midrule
      H. flip            & \multicolumn{6}{c}{\ding{51}}                                                                                                                                                               \\
      RRC                & \multicolumn{6}{c}{\ding{51}}                                                                                                                                                               \\
      Rand Augment       & {\color[HTML]{9B9B9B} \ding{55}}                                   & 3/0.5                            & 5/0.5                            & 7/0.5                            & 7/0.5 & 7/0.5 \\
      Auto Augment       & \multicolumn{6}{c}{{\color[HTML]{9B9B9B} \ding{55}}}                                                                                                                                        \\
      Mixup alpha        & 0.05                                                               & 0.1                              & 0.1                              & 0.3                              & 0.5   & 0.7   \\
      Cutmix alpha       & \multicolumn{6}{c}{1.0}                                                                                                                                                                     \\
      Erasing prob.      & \multicolumn{6}{c}{{\color[HTML]{9B9B9B} \ding{55}}}                                                                                                                                        \\
      Color Jitter       & \multicolumn{6}{c}{{\color[HTML]{9B9B9B} \ding{55}}}                                                                                                                                        \\
      PCA lighting       & \multicolumn{6}{c}{{\color[HTML]{9B9B9B} \ding{55}}}                                                                                                                                        \\
      \midrule
      SWA                & \multicolumn{6}{c}{{\color[HTML]{9B9B9B} \ding{55}}}                                                                                                                                        \\
      EMA                & \multicolumn{6}{c}{{\color[HTML]{9B9B9B} \ding{55}}}                                                                                                                                        \\
      \midrule
      Layer scale        & \multicolumn{6}{c}{{\color[HTML]{9B9B9B} \ding{55}}}                                                                                                                                        \\
      \midrule
      CE loss            & \multicolumn{6}{c}{\ding{51}}                                                                                                                                                               \\
      BCE loss           & \multicolumn{6}{c}{{\color[HTML]{9B9B9B} \ding{55}}}                                                                                                                                        \\
      \midrule
      Mixed precision    & \multicolumn{6}{c}{\ding{51}}                                                                                                                                                               \\
      \midrule
      Test crop ratio    & \multicolumn{6}{c}{0.9}                                                                                                                                                                     \\
      \midrule
      Top-1 acc. (\%)    & 73.9                                                               & 78.1                             & 80.2                             & 82.1                             & 83.1  & 83.9  \\
      \bottomrule
    \end{tabular}
  }
  \caption{ImageNet-1k training and evaluation settings for different PATNet variants.}
  \label{tab:imagenet_settings}
\end{table*}

For object detection and instance segmentation on the COCO2017 dataset, we equip our PATNet backbone with the popular Mask R-CNN detector. We use ImageNet-1k pre-trained weights to initialize the backbone and Xavier to initialize the add-on layers. Detailed settings are summarized in Table~\ref{tab:coco_settings}.

\begin{table*}[ht]
  \small
  \renewcommand{\arraystretch}{0.7}  
  \centering
  \resizebox{0.88\linewidth}{!}{%
    \setlength{\tabcolsep}{6pt}
    \begin{tabular}{@{}l|ccc@{}}
      \toprule
      Variants           & \qquad S \qquad \qquad                                            & \qquad M \qquad \qquad & L      \\
      \midrule
      Train and test Res & \multicolumn{3}{c}{shorter side $=$ 800, longer side $\leq$ 1333}                                   \\
      Batch size         & \multicolumn{3}{c}{16 (2 on each GPU)}                                                              \\
      Optimizer          & \multicolumn{3}{c}{AdamW}                                                                           \\
      Train schedule     & \multicolumn{3}{c}{1$\times$ schedule (12 epochs)}                                                  \\
      Weight decay       & \multicolumn{3}{c}{0.0001}                                                                          \\
      Warmup schedule    & \multicolumn{3}{c}{linear}                                                                          \\
      Warmup iterations  & \multicolumn{3}{c}{500}                                                                             \\
      LR decay           & \multicolumn{3}{c}{StepLR at epoch 8 and 11 with decay rate 0.1}                                    \\
      LR                 & 0.0002                                                            & 0.0001                 & 0.0001 \\
      Stoch. Depth       & 0.15                                                              & 0.2                    & 0.3    \\
      \bottomrule
    \end{tabular}%
  }
  \caption{Experimental settings of object detection and instance segmentation on the COCO2017 dataset.}
  \vspace{-0.2in}
  \label{tab:coco_settings}
\end{table*}

\subsection{Full Comparison on ImageNet-1k Benchmark.}
The full Comparison on ImageNet-1k Benchmark please refer to Table~\ref{tab:appendix class_Benchmark}, which complements the results provided in Table~1 of the main paper. 

\begin{table*}[ht] \small
  \setlength\tabcolsep{3pt}   
  \renewcommand{\arraystretch}{0.8}  
  \centering
  \begin{tabular}{@{}lcccccccc@{}}
    \hline
    Network   &\makecell{Type}   & \makecell{Params\\(M)}	&\makecell{FLOPs\\(G)}	  &\makecell{Throughput\\V100 (FPS)$\uparrow$} &\makecell{Throughput\\MI250 (FPS)$\uparrow$}	&\makecell{Latency\\CPU (ms)$\downarrow$} &\makecell{Top-1\\(\%)$\uparrow$}  \\
    \hline
    ShuffleNetV2 x1.5\cite{Ma2018}      & cnn             & 3.5              & 0.30  & 5315       & 6642        & 13.7        & 72.6       \\
    MobileNetV2\cite{Sandler2018a}      & cnn             & 3.5              & 0.31  & 3924       & 7359        & 13.7        & 72.0       \\
    FasterNet-T0\cite{Chen2023}         & cnn             & 3.9              & 0.34  & {\bf 8546} & 10612       & {\bf 10.5}  & 71.9       \\
    MobileViT-XXS\cite{Mehta2021}       & hybrid          & 1.3              & 0.42  & 2900       & 3321        & 16.7        & 69.0       \\
    MobileViTv2-0.5\cite{Mehta2022}     & hybrid          & 1.4              & 0.46  & 3094       & 3135        & 15.8        & 70.2       \\
    PATNet-T0(ours)                     & hybrid          & 4.3              & 0.25  & 7777       & {\bf 11744} & 12.2        & {\bf 73.9} \\
    \hline
    EfficientNet-B0\cite{Tan2019}       & cnn             & 5.3              & 0.39  & 2934       & 3344        & 22.7        & 77.1       \\
    GhostNet x1.3\cite{Han2020}         & cnn             & 7.4              & 0.24  & 3788       & 3620        & 16.7        & 75.7       \\
    ShuffleNetV2 x2\cite{Ma2018}        & cnn             & 7.4              & 0.59  & 4290       & 5371        & 22.6        & 74.9       \\
    MobileNetV2 x1.4\cite{Sandler2018a} & cnn             & 6.1              & 0.60  & 2615       & 4142        & 21.7        & 74.7       \\
    FasterNet-T1\cite{Chen2023}         & cnn             & 7.6              & 0.85  & {\bf 4648} & 7198        & 22.2        & 76.2       \\
    EfficientViT-B1-192\cite{Cai2023b}  & hybrid          & 9.1              & 0.38  & 4072       & 3912        & {\bf 19.3}  & 77.7       \\
    MobileViT-XS\cite{Mehta2021}        & hybrid          & 2.3              & 1.05  & 1663       & 1884        & 32.8        & 74.8       \\
    PATNet-T1(ours)                     & hybrid          & 7.8              & 0.55  & 4403       & {\bf 7379}  & 21.5        & {\bf 78.1} \\
    \hline
    EfficientNet-B1\cite{Tan2019}       & cnn             & 7.8              & 0.70  & 1730       & 1583        & 35.5        & 79.1       \\
    ResNet50\cite{He2015}               & cnn             & 25.6             & 4.11  & 1258       & 3135        & 94.8        & 78.8       \\
    FasterNet-T2\cite{Chen2023}         & cnn             & 15.0             & 1.91  & 2455       & 4189        & 43.7        & 78.9       \\
    PoolFormer-S12\cite{Yu2022a}        & hybrid          & 11.9             & 1.82  & 1927       & 3558        & 56.1        & 77.2       \\
    MobileViT-S\cite{Mehta2021}         & hybrid          & 5.6              & 2.03  & 1219       & 1370        & 52.4        & 78.4       \\
    MobileViTv2-1.0\cite{Mehta2022}     & hybrid          & 4.9              & 1.85  & 1391       & 1543        & 41.5        & 78.1       \\
    EfficientViT-B1\cite{Cai2023b}      & hybrid          & 9.1              & 0.52  & 3072       & 3387        & {\bf 25.7}  & 79.4       \\
    PATNet-T2(ours)                     & hybrid          & 12.6             & 1.03  & {\bf 3074} & {\bf 4761}  & 35.2        & {\bf 80.2} \\
    \hline
    EfficientNet-B3\cite{Tan2019}       & cnn             & 12.0             & 1.80  & 768        & 926         & 73.5        & 81.6       \\
    ConvNeXt-T\cite{Liu2022b}           & cnn             & 28.6             & 4.47  & 902        & 1103        & 99.4        & {\bf 82.1} \\
    FasterNet-S\cite{Chen2023}          & cnn             & 31.1             & 4.56  & 1261       & 2243        & 96.0        & 81.3       \\
    PoolFormer-S36\cite{Yu2022a}        & hybrid          & 30.9             & 5.00  & 675        & 1092        & 152.4       & 81.4       \\
    MobileViTv2-1.5\cite{Mehta2022}     & hybrid          & 10.6             & 4.00  & 812        & 1000        & 104.4       & 80.4       \\
    MobileViTv2-2.0\cite{Mehta2022}     & hybrid          & 18.5             & 7.50  & 551        & 684         & 103.7       & 81.2       \\
    Swin-T\cite{Liu2021c}               & hybrid          & 28.3             & 4.51  & 808        & 1192        & 107.1       & 81.3       \\
    PATNet-S(ours)                      & hybrid          & 29.0             & 2.71  & {\bf 1559} & {\bf 2422}  & {\bf 72.5}  & {\bf 82.1} \\
    \hline
    EfficientNet-B4\cite{Tan2019}       & cnn             & 19.0             & 4.20  & 356        & 442         & 156.9       & 82.9       \\
    ConvNeXt-S\cite{Liu2022b}           & cnn             & 50.2             & 8.71  & 510        & 610         & 185.5       & {\bf 83.1} \\
    FasterNet-M\cite{Chen2023}          & cnn             & 53.5             & 8.74  & 621        & 1098        & 181.6       & 83.0       \\
    PoolFormer-M36\cite{Yu2022a}        & hybrid          & 56.2             & 8.80  & 444        & 721         & 244.3       & 82.1       \\
    Swin-S\cite{Liu2021c}               & hybrid          & 49.6             & 8.77  & 477        & 732         & 199.1       & 83.0       \\
    PATNet-M(ours)                      & hybrid          & 61.3             & 6.69  & {\bf 799}  & {\bf 1280}  & {\bf 155.3} & {\bf 83.1} \\
    \hline
    EfficientNet-B5\cite{Tan2019}       & cnn             & 30.0             & 9.90  & 246        & 313         & 333.3       & 83.6       \\
    ConvNeXt-B\cite{Liu2022b}           & cnn             & 88.6             & 15.38 & 322        & 430         & 317.1       & 83.8       \\
    FasterNet-L\cite{Chen2023}          & cnn             & 93.5             & 15.52 & 384        & 709         & 312.5       & 83.5       \\
    PoolFormer-M48\cite{Yu2022a}        & hybrid          & 73.5             & 11.59 & 335        & 556         & 322.3       & 82.5       \\
    Swin-B\cite{Liu2021c}               & hybrid          & 87.8             & 15.47 & 315        & 520         & 333.8       & 83.5       \\
    PATNet-L(ours)                      & hybrid          & 104.3            & 11.91 & {\bf 426}  & {\bf 765}   & {\bf 272.5} & {\bf 83.9} \\
    \hline
  \end{tabular}
  \caption{Full comparison on ImageNet-1k Benchmark: models with similar top-1 accuracy are grouped together. The best results are in bold.}
  \label{tab:appendix class_Benchmark}
\end{table*}

\subsection{Ablation Studies}
\textbf{Partial Visual Attention vs. Conventional Visual Attention.}
To further prove the superiority of our PAT, we present experiment results for the combination of our partial attention and classic visual attention networks, and the results are shown in Table~\ref{tab:abla_convention_attention_acc}. The results demonstrate the effectiveness of our enhanced Gaussian-SE module.
\begin{table}[ht]
  \setlength\tabcolsep{3.8pt}   
   \renewcommand{\arraystretch}{0.8}  
  \centering
  \begin{tabular}{@{}lcccc c@{}}
    \hline
    Visual type & \makecell{Params(M)} & \makecell{FLOPs(G)} & \makecell{Throughput(fps)$\uparrow$} & \makecell{latency(ms)$\downarrow$} & \makecell{Acc1(\%)$\uparrow$} \\
    \hline
    SRM~\cite{Lee2019a}               & 12.2       & 1.03     & 4751         & 35.2        & 79.6                          \\
    SE-NET~\cite{Hu2018}              & 12.3       & 1.04     & 4910         & 32.3        & 79.8                          \\
    PAT(ours)                         & 12.6       & 1.03     & 4761         & 35.2        & {\bf 80.2}                     \\
    \hline
  \end{tabular}
  \caption{Comparison on PATNet-T2 of partial visual attention and conventional visual attention on ImageNet1K dataset.}
  \label{tab:abla_convention_attention_acc}
\end{table}

\textbf{Comparison On ImageNet-1k Under Same Training Settings.}
In order to further verify the effectiveness and fair comparison of our PATNet, we reproduce the results of FastNet on ImageNet-1k but based on our training experiment configuration, the results are shown in Table~\ref{tab:pat_vs_fasternet}. It can be seen from the results that our PATNet still has great advantages.

\begin{table*}[ht] \small
  \setlength\tabcolsep{3pt}   
  \centering
  \begin{tabular}{@{}lcccccccc@{}}
    \hline
    Network   &\makecell{Type}   & \makecell{Params\\(M)}	&\makecell{FLOPs\\(G)}	  &\makecell{Throughput\\V100 (FPS)$\uparrow$} &\makecell{Throughput\\MI250 (FPS)$\uparrow$}	&\makecell{Latency\\CPU (ms)$\downarrow$} &\makecell{Top-1\\(\%)$\uparrow$}  \\
    \hline
    FasterNet-T0\cite{Chen2023}         & cnn             & 3.9              & 0.34  & {\bf 8546} & 10612       & {\bf 10.5}  & 71.9       \\
    FasterNet-T0*\cite{Chen2023}        & cnn             & 3.9              & 0.34  & {\bf 8546} & 10612       & {\bf 10.5}  & 71.0       \\
    PATNet-T0(ours)                     & hybrid          & 4.3              & 0.25  & 7777       & {\bf 11744} & 12.2        & {\bf 73.9} \\
    \hline
    FasterNet-T1\cite{Chen2023}         & cnn             & 7.6              & 0.85  & {\bf 4648} & 7198        & 22.2        & 76.2       \\
    FasterNet-T1*\cite{Chen2023}        & cnn             & 7.6              & 0.85  & {\bf 4648} & 7198        & 22.2        & 76.5       \\
    PATNet-T1(ours)                     & hybrid          & 7.8              & 0.55  & 4403       & {\bf 7379}  & 21.5        & {\bf 78.1} \\
    \hline
    FasterNet-T2\cite{Chen2023}         & cnn             & 15.0             & 1.91  & 2455       & 4189        & 43.7        & 78.9       \\
    FasterNet-T2*\cite{Chen2023}        & cnn             & 15.0             & 1.91  & 2455       & 4189        & 43.7        & 79.2       \\
    PATNet-T2(ours)                     & hybrid          & 12.6             & 1.03  & {\bf 3074} & {\bf 4761}  & 35.2        & {\bf 80.2} \\
    \hline
    \hline
    FasterNet-S\cite{Chen2023}          & cnn             & 31.1             & 4.56  & 1261       & 2243        & 96.0        & 81.3       \\
    FasterNet-S\cite{Chen2023}          & cnn             & 31.1             & 4.56  & 1261       & 2243        & 96.0        & 81.5       \\
    PATNet-S(ours)                      & hybrid          & 29.0             & 2.71  & {\bf 1559} & {\bf 2422}  & {\bf 72.5}  & {\bf 82.1} \\
    \hline
    FasterNet-M\cite{Chen2023}          & cnn             & 53.5             & 8.74  & 621        & 1098        & 181.6       & 83.0       \\
    FasterNet-M*\cite{Chen2023}         & cnn             & 53.5             & 8.74  & 621        & 1098        & 181.6       & 83.0       \\
    PATNet-M(ours)                      & hybrid          & 61.3             & 6.69  & {\bf 799}  & {\bf 1280}  & {\bf 155.3} & {\bf 83.1} \\
    \hline
    FasterNet-L\cite{Chen2023}          & cnn             & 93.5             & 15.52 & 384        & 709         & 312.5       & 83.5       \\
    FasterNet-L*\cite{Chen2023}         & cnn             & 93.5             & 15.52 & 384        & 709         & 312.5       & 83.6       \\
    PATNet-L(ours)                      & hybrid          & 104.3            & 11.91 & {\bf 426}  & {\bf 765}   & {\bf 272.5} & {\bf 83.9} \\
    \hline
  \end{tabular}
  \caption{Comparison on ImageNet-1k. The "*" denotes reproduction results based on our experimental setup.}
  \label{tab:pat_vs_fasternet}
\end{table*}

\clearpage
\setcounter{page}{1}
\bibliography{egbib}

\end{document}